\documentclass[letterpaper, 10 pt, conference]{ieeeconf}

\IEEEoverridecommandlockouts
\usepackage{cite}
\usepackage{amsmath}
\usepackage{amssymb}
\usepackage{graphicx}
\usepackage{booktabs}
\usepackage{xspace}
\usepackage{url}
\usepackage{hyperref}
\usepackage{color}
\usepackage{multirow}
\usepackage{siunitx}

\newcommand{\etal}{et al.\xspace}

\DeclareMathOperator*{\argmin}{arg\,min}

\let\origSigma\Sigma
\renewcommand{\Sigma}{\boldsymbol{\origSigma}}
\newcommand{\R}{\mathbb{R}}

\title{\LARGE \bf
Delay-Aware Active Triangulation with Uncertainty-Driven\\Multi-Agent Reinforcement Learning for Counter-UAS
}
\author{Seungwook Lee$^{1}$, David Hyunchul Shim$^{1}$
\thanks{$^{1}$S. Lee and D. H. Shim are with the Korea Advanced Institute of Science and Technology (KAIST), Daejeon, Korea. Email: \{seungwook1024, hcshim\}@kaist.ac.kr}%
}

\begin{document}

\maketitle
\thispagestyle{empty}
\pagestyle{empty}

\begin{abstract}
Multi-agent active visual triangulation enables precise 3D localization of aerial targets by coordinating mobile observers with controllable cameras. However, existing methods assume instantaneous state feedback, ignoring cumulative latency from detection, communication, and decision propagation. We present a delay-aware, uncertainty-driven multi-agent reinforcement learning framework for target localization in Counter-UAS applications. Our contributions are: (1)~a Dec-POMDP formulation with Age-of-Information (AoI) augmented observations enabling delay-aware coordination---AoI improves triangulation validity by 10.6 percentage points; (2)~a controlled comparison showing that perception-consistent rewards outperform privileged clean-state rewards ($0.547$\,m vs.\ $0.633$\,m RMSE, 27\% fewer track losses)---both policies are trained through identical observation noise but differ in what they are optimized for, producing a stability--robustness tradeoff; and (3)~multi-source analytical covariance propagation incorporating pixel, pose, gimbal, and intrinsics uncertainties---restricting to angular noise alone causes 2.8-fold RMSE degradation. Experiments with MAPPO in 4096 parallel environments achieve $0.547 \pm 0.217$\,m RMSE with $78.1\%$ triangulation validity, while MLP policies achieve near-zero validity ($0.7\%$), confirming recurrent memory as essential for delay compensation.
\end{abstract}

\section{Introduction}
\label{sec:introduction}

The proliferation of small unmanned aerial systems (UAS) has created urgent demand for Counter-UAS (C-UAS) technologies capable of detecting, tracking, and neutralizing aerial threats~\cite{Barisic2022SSRR,Castrillo2022}. Regardless of the neutralization method, effective C-UAS response depends on precise, continuous 3D target localization. Vision-based tracking offers passive operation, rich semantics, and scalable deployment on small platforms.

Multi-agent \emph{active triangulation}---where mobile observers dynamically reposition to optimize measurement geometry---provides instantaneous 3D position estimates without the convergence delays of bearing-only filtering based methods~\cite{Hartley2013}. Modern platforms further enable controllable sensors: gimbal-stabilized cameras with optical zoom decouples the viewing direction from platform motion and the distance from the subject.

Despite these capabilities, real-world multi-agent systems face a critical bottleneck: \emph{system delays}. Cumulative latency from frame acquisition, object detection, inter-agent communication, and control propagation can range from tens to thousands of milliseconds. These delays are asymmetric: ego-state estimates benefit from high-rate onboard sensors and computations, whereas inter-agent observations traverse the full delay pipeline. Prior perception-aware methods are predominantly single-agent~\cite{Falanga2018} and do not encounter this multi-agent delay structure.

Gavin~\etal~\cite{Gavin2024} demonstrated multi-agent RL for triangulation with analytical covariance-based rewards and real world flight tests. We extend their framework to incorporate realistic deployment conditions: stochastic communication delays with perspective asymetric staleness, controllable gimbal-zoom sensors with field-of-view constraints, and multi-source uncertainty propagation. Our three contributions are:

\begin{enumerate}
\item \textbf{Delay-aware Dec-POMDP formulation for active triangulation}: We show that multi-agent active triangulation requires explicit delay modeling under realistic communication conditions. Formulating the problem as a Dec-POMDP with AoI-augmented observations~\cite{fu2025rainbow,wang2024addressing} and recurrent policies, AoI improves triangulation validity by 10.6\,\%p, while MLP policies with stacked observations achieve near-zero validity ($0.7\%$), confirming that recurrence and staleness metadata are both necessary.

\item \textbf{Comparative evaluation of privileged and perception-consistent rewards}: Following the privileged-information paradigm~\cite{pmlr-v180-baisero22a}, we implement a dual-path delay architecture and conduct a controlled comparison. Both policies are trained through identical observation noise but differ in what they are optimized for: perception-consistent rewards outperform in aggregate metrics ($0.547$\,m vs.\ $0.633$\,m RMSE, 27\% fewer track losses, 38\% fewer collisions), while per-timestep analysis reveals a stability--robustness tradeoff where privileged rewards yield more temporally stable coordination. Neither formulation strictly dominates, identifying reward--observation alignment as a deployment-dependent design axis.

\item \textbf{Multi-source analytical covariance propagation}: We extend Gavin~\etal's angular-noise covariance~\cite{Gavin2024} to incorporate pixel detection noise, pose error, gimbal calibration, and camera intrinsics uncertainties. Multi-source propagation improves triangulation validity from $32.2\%$ to $78.1\%$ with 2.8-fold RMSE reduction over angular-only modeling.
\end{enumerate}

\begin{figure*}[!t]
\vspace{2mm}
\centering
\begin{minipage}[t]{0.25\textwidth}
\centering
\includegraphics[width=\textwidth]{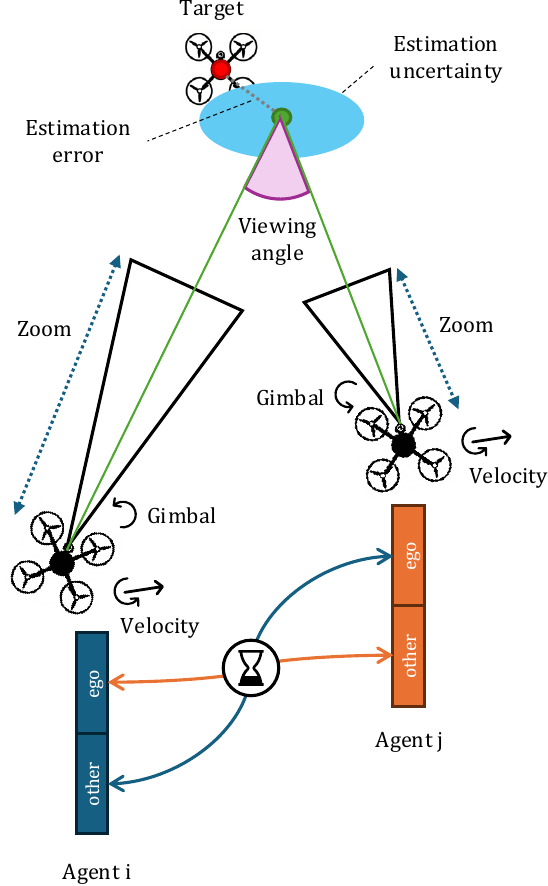}
\end{minipage}\hfill
\begin{minipage}[t]{0.65\textwidth}
\centering
\includegraphics[width=\textwidth]{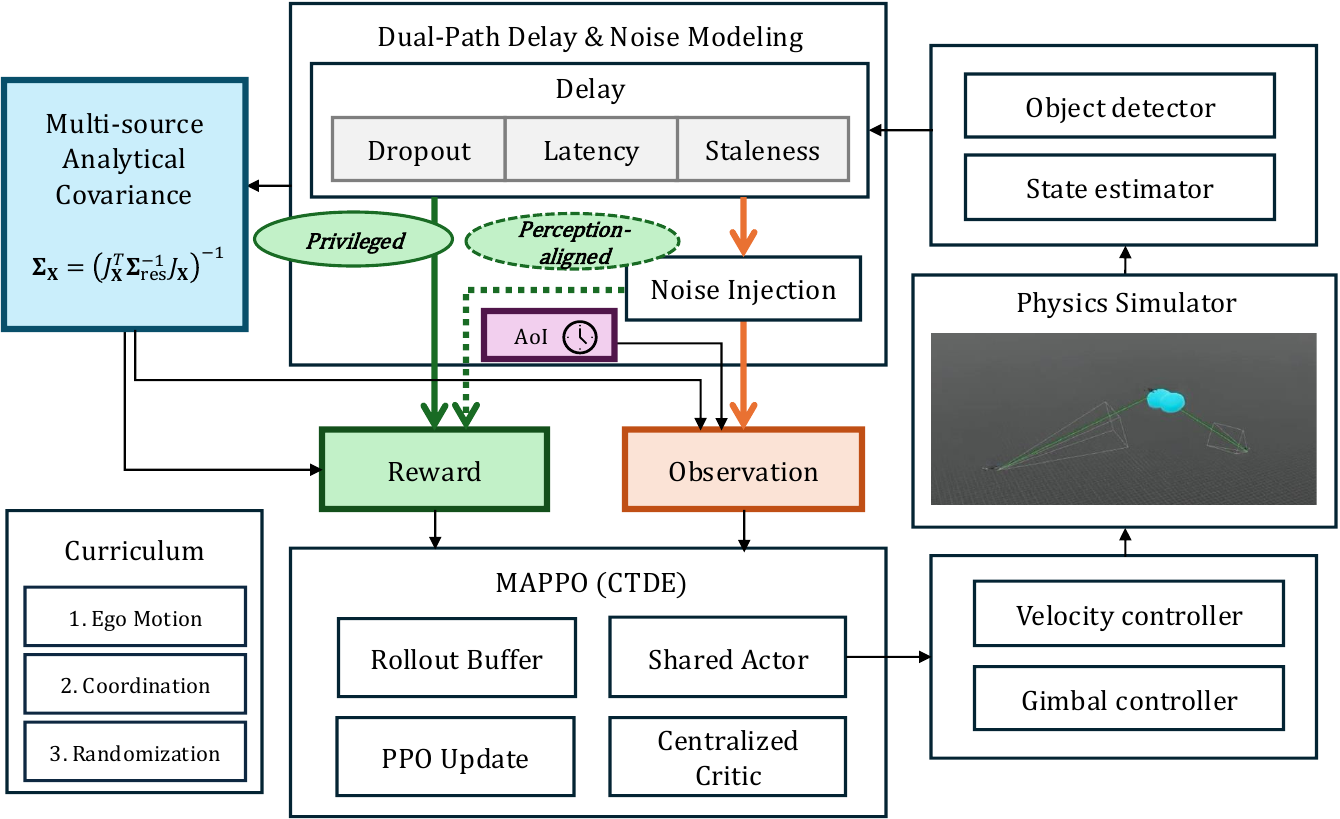}
\end{minipage}
\caption{\textbf{Left}: Problem setup showing two agents with gimbaled zoom cameras coordinating to localize a target using triangulation. Each agent maintains ego and other observation streams subject to asymmetric delays, with estimation uncertainty and error indicated. \textbf{Right}: Training architecture with the dual-path delay and noise modeling pipeline. The dual-path architecture separates the observation path (always noisy and delayed) from the reward path: the privileged reward (green, solid) computes rewards from clean delayed states, while the added perception-consistent reward (green, dashed) uses the same noisy delayed states as the observation. Both configurations train under MAPPO with a three-phase curriculum.}
\label{fig:system-architecture}
\end{figure*}

\section{Related Work}
\label{sec:related-work}

\textbf{C-UAS and active perception.}
Counter-UAS research spans multi-robot tracking~\cite{Barisic2022SSRR}, swarm defense~\cite{Brust2021}, and cooperative methodologies~\cite{Castrillo2022}. LiDAR-based systems achieve high precision at short range~\cite{Vrba2025TROCUASLidarSOTA}, while vision-based approaches offer lighter, longer-range alternatives: Saviolo~\etal~\cite{Saviolo2025} demonstrate single-agent tracking exceeding 50\,km/h using adaptive zoom. Perception-aware MPC~\cite{Falanga2018} and information-theoretic guidance address single-agent visibility planning but do not model stochastic inter-agent communication delays inherent in distributed systems.

\textbf{Multi-view geometry and active triangulation.}
Classical multi-view geometry~\cite{Hartley2013} and stereo error modeling~\cite{MatthiesShafer1987} underpin modern uncertainty propagation. Recent work further characterizes triangulation uncertainty under realistic sensing conditions: DiLeo~\etal~\cite{DiLeo2010} analyze covariance propagation through stereo geometry and Lee~\etal~\cite{LeeCivera2020} develop robust uncertainty-aware multiview triangulation with outlier rejection. Most directly related, Gavin~\etal~\cite{Gavin2024} pioneered analytical covariance-based rewards for learned multi-agent triangulation with real-flight demonstrations. We extend their framework to handle stochastic communication delays, controllable gimbal-zoom sensors with FoV constraints, and multi-source uncertainty propagation through the full observation pipeline.

\textbf{Delay-aware RL and AoI.}
Wang~\etal~\cite{wang2024addressing} formalized Delayed-Observation MDPs showing standard RL suffers under signal delays. Fu~\etal~\cite{fu2025rainbow} introduced the DSID-POMDP for stochastic per-component delays in multi-agent settings. Age-of-Information~\cite{10589619,Kaul2012AoI}, quantifies data freshness and has been applied to multi-agent scheduling~\cite{ROY2024101141}. Baisero~\etal~\cite{pmlr-v180-baisero22a} formalized asymmetric training with privileged information for POMDPs. MAPPO~\cite{Yu2022} provides effective Centralized Training Decentralized Execution(CTDE) framework for cooperative tasks. We utilize AoI-augmented observations and a dual-path stochastic delay modeling architecture to enable delay-aware coordination and conduct a controlled comparison of privileged vs.\ perception-consistent rewards.

\begin{figure}[!t]
\centering
\includegraphics[width=\columnwidth]{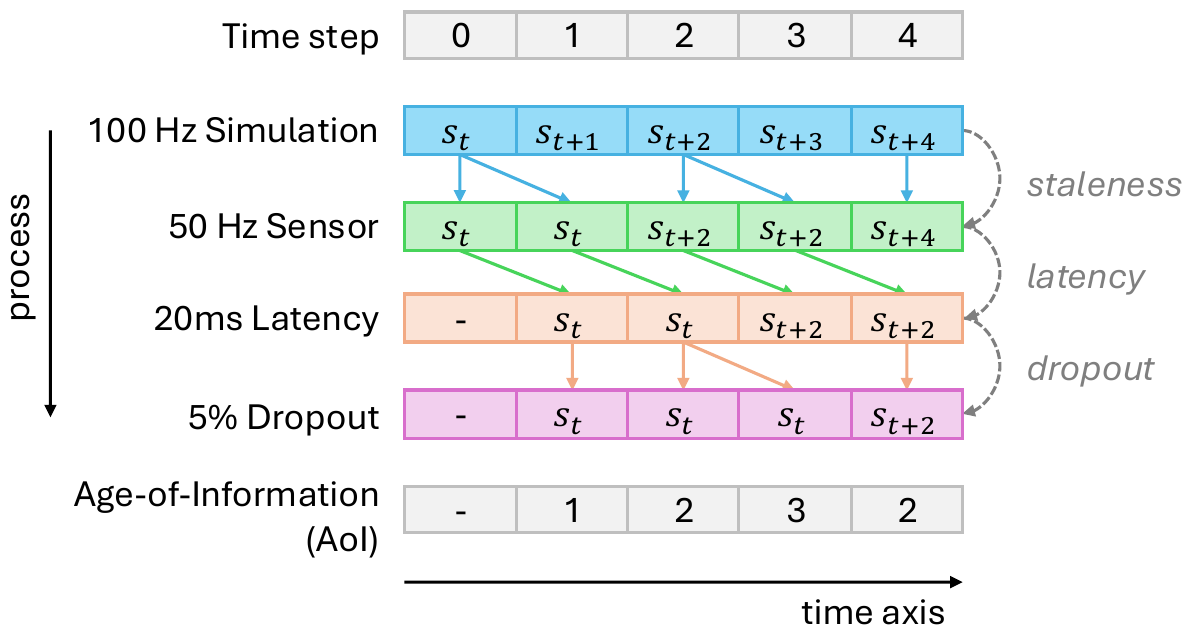}
\caption{Three-stage delay pipeline: staleness (sample-and-hold), latency (buffered delay), and dropout (packet loss). This logic is used to model realistic delays in the training environment. An identical delay structure is applied to each agent's ego-state stream and inter-agent state streams. The ego-state stream has reduced delay parameters to reflect near-zero latency from onboard sensors, while inter-agent streams have higher delay parameters matching realistic communication latencies.}
\label{fig:delay-pipeline}
\end{figure}

\section{Problem Formulation}
\label{sec:problem-formulation}

We formulate multi-agent active visual triangulation as a Dec-POMDP with stochastic communication delays.

\subsection{Dec-POMDP with Delayed Observations}
\label{sec:dec-pomdp}

Consider $C$ agents (camera-equipped drones) indexed by $i \in \mathcal{I}=\{1,\dots,C\}$ and a single target $T$. The system operates in discrete steps $k = 0, 1, \dots, K{-}1$ with step duration $\Delta t$. We model the problem as a Dec-POMDP tuple $\langle \mathcal{I}, \mathcal{S}, \{\mathcal{A}^i\}_{i \in \mathcal{I}}, P, R, \{\Omega^i\}_{i \in \mathcal{I}}, Z, \gamma \rangle$ augmented with a stochastic communication model, where $\mathcal{S}$ is the global state space, $\mathcal{A}^i$ is agent $i$'s action space, $P$ is the state-transition model, $R$ is the shared team reward, $\Omega^i$ is agent $i$'s observation space, $Z$ is the observation model induced by sensing and communication, and $\gamma$ is the discount factor. The global state $\mathbf{s}_k = (\mathbf{x}_k^1, \dots, \mathbf{x}_k^C, \mathbf{x}_k^T)$ comprises each agent state $\mathbf{x}_k^i \in \R^{16}$ (3D position, 3D velocity, 4D quaternion orientation, 3D angular velocity, 2D gimbal angles, 1D zoom) and target state $\mathbf{x}_k^T \in \R^6$ (3D position, 3D velocity).

Each agent executes a 7-dimensional continuous action:
\begin{equation}
\label{eq:action}
a^{(i)}_k = \left[a_{v_x}, a_{v_y}, a_{v_z}, a_{\dot{\psi}}, a_{\dot{\alpha}}, a_{\dot{\beta}}, a_{\dot{z}}\right]^\top \in [-1, 1]^7
\end{equation}
mapped to physical commands via per-dimension scaling: 3D velocity, yaw rate, gimbal pan/tilt rates, and zoom rate.

We assume all states are subject to the stochastic delay model consisting of \emph{sampling interval}, \emph{latency}, and \emph{dropouts}. For inter-agent communication, agents obtain teammate states through a stochastic communication channel with three delay sources: random inter-transmission intervals, transport latency, and packet dropout. For self state observation, each agent observes its own state through the same process with reduced delay (detailed in Section~\ref{sec:delay-system}). Each agent~$i$ maintains a held record $\tilde{\mathbf{x}}^{(j \to i)}_k$ of the latest received state from neighbor $j \in \mathcal{N}_i=\{1,\dots,C\} \setminus \{i\}$, updated only upon successful reception:
\begin{equation}
\label{eq:sample-hold}
\tilde{\mathbf{x}}^{(j \to i)}_{k+1} = \begin{cases}
\mathbf{x}^{(j \to i)}_{k+1 - d_{k+1}^{(j \to i)}} & \text{if } \chi^{(j \to i)}_{k+1} = 1 \\
\tilde{\mathbf{x}}^{(j \to i)}_k & \text{if } \chi^{(j \to i)}_{k+1} = 0
\end{cases}
\end{equation}
where $\chi^{(j \to i)}_k \in \{0,1\}$ indicates successful reception and $d_k^{(j \to i)}$ is the end-to-end delay.

We incorporate an Age-of-Information (AoI) tag~\cite{Kaul2012AoI} $\Delta^{(j \to i)}_k$ for each inter-agent observation, measuring elapsed time since the source generated the currently held information:
\begin{equation}
\label{eq:aoi}
\Delta^{(j \to i)}_{k+1} = \begin{cases}
d_{k+1}^{(j \to i)} & \text{if } \chi^{(j \to i)}_{k+1} = 1 \\
\Delta^{(j \to i)}_k + 1 & \text{if } \chi^{(j \to i)}_{k+1} = 0
\end{cases}
\end{equation}
Let $\tilde{\mathcal{X}}^j$ denote the space of delayed, noisy records of agent $j$'s state. Agent $i$'s observation space is
\begin{equation}
\label{eq:observation-space}
\Omega^i = \tilde{\mathcal{X}}^i \times \prod_{j \in \mathcal{N}_i} \left( \tilde{\mathcal{X}}^j \times \mathbb{R}_{\ge 0} \right),
\end{equation}
and the AoI-augmented observation for agent $i$ is:
\begin{equation}
\label{eq:observation}
o^i_k = \Big( \tilde{\mathbf{x}}^i_k, \; \{\tilde{\mathbf{x}}^{(j \to i)}_k : j \in \mathcal{N}_i\}, \; \{\Delta^{(j \to i)}_k : j \in \mathcal{N}_i\} \Big)
\end{equation}
Including AoI enables policies to reason about information freshness---discounting stale teammate information for coordination decisions. Unlike prior delay-aware formulations that rely on assumed delay models or latent-delay compensation, we expose AoI as a directly measured per-message signal, computed as current simulator time minus the message timestamp. To account for delayed and asynchronously updated teammate observations, we use a recurrent policy (Gated Recurrent Unit, GRU) that maintains a temporal belief over recent information history. This extends the DSID-POMDP framework~\cite{fu2025rainbow} to multi-robot applications with synchronized clocks, where a key feature is the asymmetry between smaller delays for ego-state observations and larger delays for inter-agent observations.
The stochastic delay process therefore augments the standard Dec-POMDP observation model $Z$ by mapping the current state and joint action to held, timestamped local records rather than instantaneous state observations.

\subsection{Triangulation Geometry}
\label{sec:triangulation-geometry}

Each agent carries a gimbaled camera with pinhole projection and linear optical zoom. For a 3D point $\mathbf{X}_c = [X, Y, Z]^\top$ in camera frame, the projection yields:
\begin{equation}
\label{eq:projection}
u = f_x \frac{X}{Z} + c_x, \quad v = f_y \frac{Y}{Z} + c_y
\end{equation}
where focal lengths scale linearly with zoom factor $z \geq 1$ as $f_x^{(z)} = z \cdot f_{x,0}$.

Given $C$ cameras with positions $\mathbf{p}_i$ and unit ray directions $\hat{\mathbf{d}}_i$ from unprojecting detected pixel coordinates, the target position minimizes the sum of squared perpendicular distances to all rays~\cite{Hartley2013}:
\begin{equation}
\label{eq:triangulation}
\hat{\mathbf{X}} = \argmin_{\mathbf{X}} \sum_{i=1}^{C} \left\| \left(I - \hat{\mathbf{d}}_i \hat{\mathbf{d}}_i^\top\right) (\mathbf{X} - \mathbf{p}_i) \right\|^2
\end{equation}
With $P_i = I - \hat{\mathbf{d}}_i \hat{\mathbf{d}}_i^\top$, this yields the closed-form solution:
\begin{equation}
\label{eq:triangulation-solution}
\hat{\mathbf{X}} = \left(\sum_{i=1}^{C} P_i\right)^{-1} \left(\sum_{i=1}^{C} P_i \, \mathbf{p}_i\right)
\end{equation}
requiring at least two cameras with non-parallel rays. We adopt this closed-form multi-view ray-intersection formulation for computational efficiency and stable integration into per-step reward evaluation.

\subsection{Objectives}
\label{sec:objectives}

The goal is to learn a joint policy $\pi = (\pi^1, \dots, \pi^C)$ maximizing:
\begin{equation}
\label{eq:objective}
\max_{\pi} \; \mathbb{E} \left[ \sum_{k=0}^{K-1} \gamma^k \sum_{i=1}^{C} r_k^{(i)} \right]
\end{equation}
where the per-agent reward decomposes into: 

\noindent\textbf{(i) Triangulation quality.} This term drives agents toward configurations that minimize covariance $\Sigma_{\mathbf{X}}$ (Section~\ref{sec:analytical-cov}):
\begin{equation}
\label{eq:reward-tri}
r_\text{tri} = \begin{cases} \dfrac{s_\text{tri}}{\sqrt{\mathrm{tr}(\Sigma_{\mathbf{X}})}} & \text{if } \hat{\mathbf{X}} \text{ exists} \\[6pt] 0 & \text{otherwise} \end{cases}
\end{equation}

\noindent\textbf{(ii) Image-space tracking.} This term encourages target centering and an appropriate bounding-box size:
\begin{equation}
\label{eq:reward-image}
r_\text{img} = s_\text{ctr} \cdot \exp\!\big({-}\lambda_\text{ctr} \|\mathbf{c} - \mathbf{c}_0\|\big) + s_\text{sz} \cdot \exp\!\big({-}\lambda_\text{sz} |A - A_\text{ref}|\big)
\end{equation}
where $\mathbf{c}$ and $\mathbf{c}_0$ are the detected and image-center bounding-box positions (normalized), and $A$, $A_\text{ref}$ are the detected and reference bounding-box areas.

\noindent\textbf{(iii) Safety constraints.} A collision penalty activates when pairwise agent distances fall below $d_\text{min}$:
\begin{equation}
\label{eq:collision}
r_\text{coll}^{(i)} = -s_\text{coll} \cdot \Delta t \cdot \sum_{j \neq i} \mathbf{1}[\|\mathbf{p}_i - \mathbf{p}_j\| < d_\text{min}]
\end{equation}

\noindent\textbf{(iv) Action regularization.} This term penalizes action magnitude and jerk.

The total per-agent reward is:
\begin{equation}
\label{eq:reward-total}
r_k^{(i)} = r_\text{tri} + r_\text{img} + r_\text{coll} + r_\text{act}.
\end{equation}

\begin{table}[!t]
\centering
\small
\caption{Evaluation metrics used in Section~\ref{sec:experiments}.}
\label{tab:metric-definitions}
\setlength{\tabcolsep}{3pt}
\begin{tabular}{p{0.20\columnwidth} p{0.12\columnwidth} p{0.6\columnwidth}}
\toprule
\textbf{Metric} & \textbf{Unit} & \textbf{Definition} \\
\midrule
RMSE & m & Euclidean error between triangulated position and ground truth. \\
\midrule
Tri Valid & \% & Fraction of steps with $\geq 2$ valid detections and non-degenerate geometry. \\
\midrule
Visibility & \% & Fraction of steps with $\geq 2$ valid detections. \\
\midrule
Convergence & steps & Steps to first valid triangulation. \\
\midrule
Collisions & \% & Collision events per episode normalized by the episode length and expressed as a percentage. \\
\midrule
Track Loss & count/ep. & Valid-to-invalid triangulation transitions per episode. \\
\bottomrule
\end{tabular}
\setlength{\tabcolsep}{6pt}
\end{table}

\begin{table*}[!t]
\vspace{3mm}
\caption{Ablation results.}
\label{tab:ablation-results}
\centering
\small
\begin{tabular}{l r l l l r r r}
\toprule
\textbf{Method} & \textbf{$N$} & \textbf{RMSE (m)} & \textbf{Tri Valid (\%)} & \textbf{Visibility (\%)} & \textbf{Conv. (steps)} & \textbf{Coll. (\%)} & \textbf{Track Loss} \\
\midrule
Noisy reward      & 4096 & $\mathbf{0.547}$$\,\pm\,$\scriptsize{0.217} & $78.1$$\,\pm\,$\scriptsize{10.3} & $84.1$$\,\pm\,$\scriptsize{6.3}           & 75.3          & $\mathbf{0.155}$ & $\mathbf{6.1}$ \\
Clean reward      & 4412 & $0.633$$\,\pm\,$\scriptsize{0.345}          & $\mathbf{79.1}$$\,\pm\,$\scriptsize{17.4}          & $\mathbf{86.7}$$\,\pm\,$\scriptsize{11.0} & $\mathbf{51.9}$ & $0.251$          & $8.3$ \\
No AoI            & 4280 & $0.701$$\,\pm\,$\scriptsize{0.360}          & $68.5$$\,\pm\,$\scriptsize{29.9}          & $83.2$$\,\pm\,$\scriptsize{13.9}          & 52.0          & $0.269$          & $8.9$ \\
Angular-only cov  & 4096 & $1.546$$\,\pm\,$\scriptsize{1.134}          & $32.2$$\,\pm\,$\scriptsize{24.2}          & $54.9$$\,\pm\,$\scriptsize{16.1}          & 89.1          & $0.193$          & $17.3$ \\
MLP (stacked obs) & 6784 & $1.561$$\,\pm\,$\scriptsize{2.495}          & $0.7$$\,\pm\,$\scriptsize{1.3}            & $2.2$$\,\pm\,$\scriptsize{2.5}            & 5.0           & $0.260$          & $0.9$ \\
\bottomrule
\end{tabular}
\end{table*}

\begin{figure*}[!t]
\vspace{2mm}
\centering
\includegraphics[width=\textwidth]{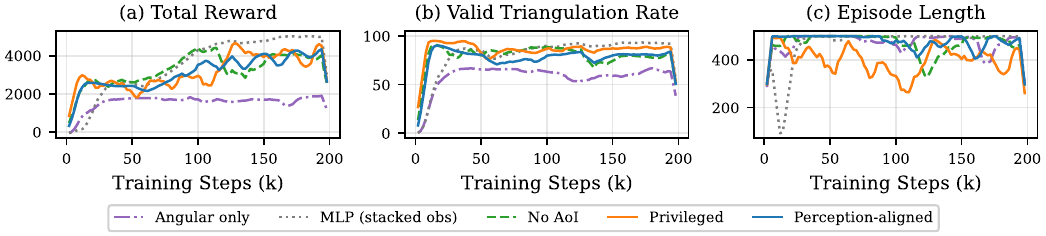}
\caption{Training curves over 200k agent steps. (a)~Total reward, (b)~valid triangulation rate, (c)~episode length before termination. Experimental results from Section~\ref{sec:experiments} show the training metrics don't necessarily reflect policy performance in terms of downstream metrics like RMSE and valid triangulation rate.}
\label{fig:training-curves}
\end{figure*}

\begin{figure*}[!t]
\centering
\begin{minipage}[t]{0.48\textwidth}
\centering
\includegraphics[width=\textwidth]{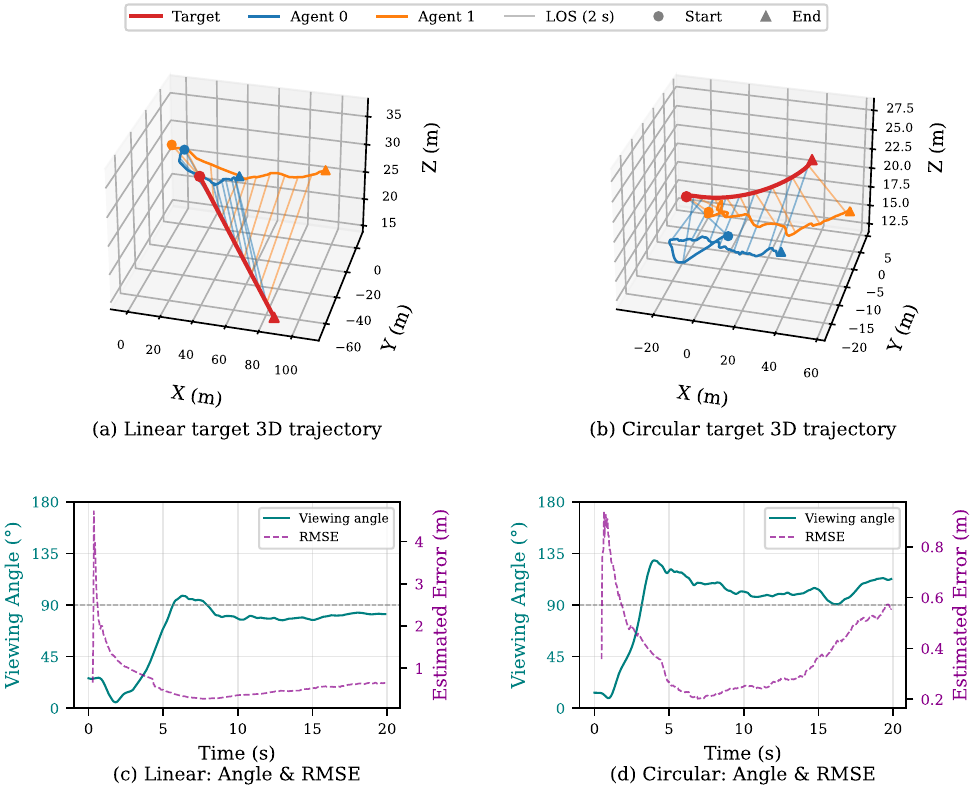}
\end{minipage}\hfill
\begin{minipage}[t]{0.48\textwidth}
\centering
\includegraphics[width=\textwidth]{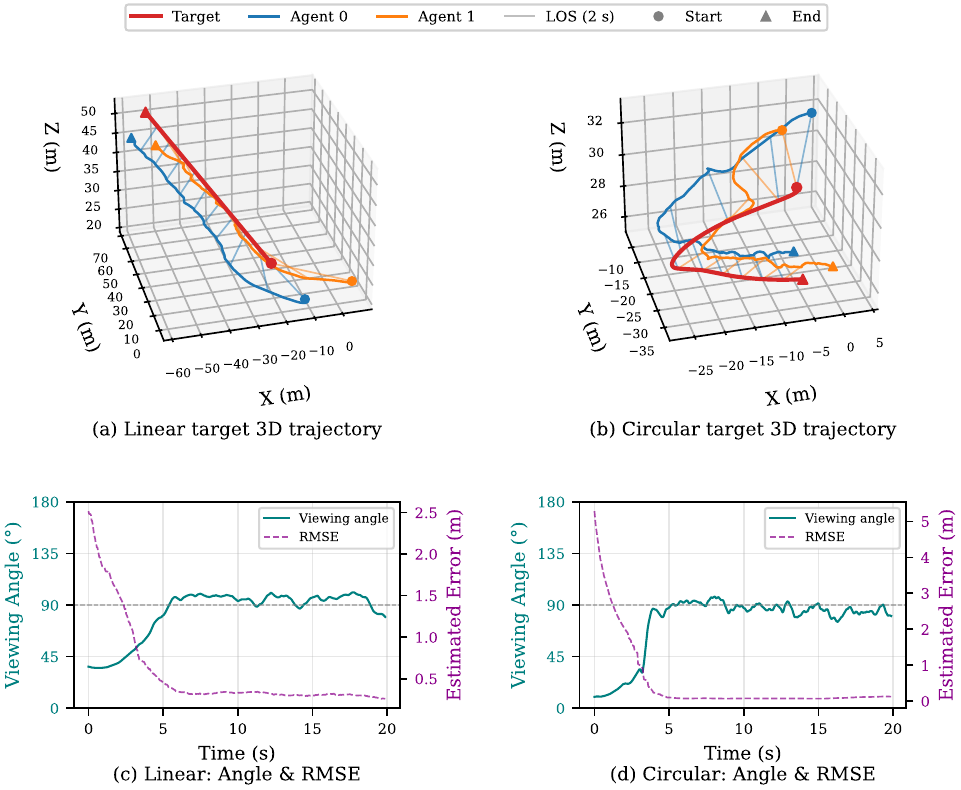}
\end{minipage}
\caption{Representative 3D trajectories of 2 agents and one target. \textbf{Left}: Perception-consistent (noisy reward). \textbf{Right}: Privileged (clean reward). The noisy reward policy initially overshoots then adjusts gently, while the clean-reward policy converges to a tighter formation with more frequent adjustments.}
\label{fig:trajectories}
\end{figure*}

\begin{figure*}[!t]
\centering
\includegraphics[width=\textwidth]{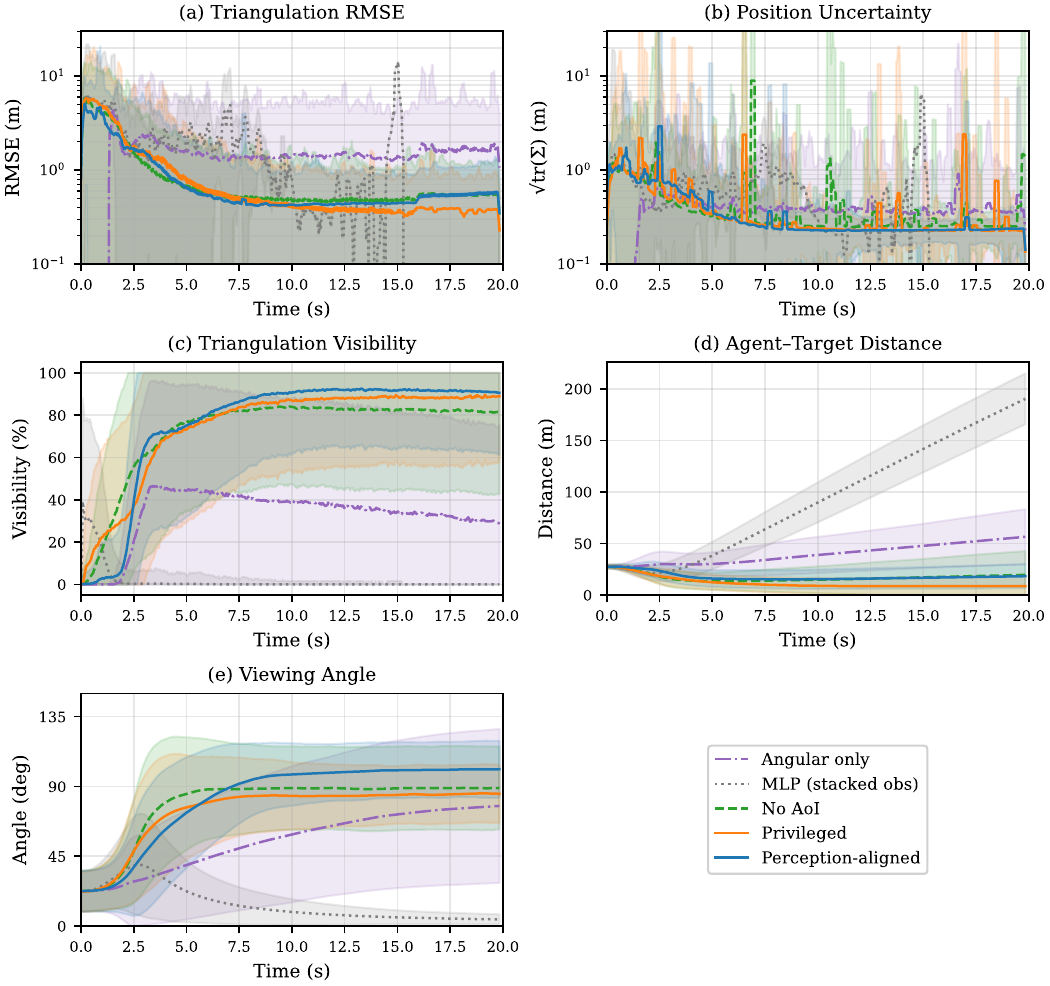}
\caption{Per-timestep metrics over 20\,s episodes, averaged across $N > 4{,}000$ episodes. (a)~RMSE, (b)~estimation uncertainty $\sqrt{\mathrm{tr}(\Sigma)}$, (c)~visibility, (d)~agent--target distance. The noisy-reward policy achieves lower RMSE but exhibits gradual temporal drift, while the clean-reward policy maintains more stable metrics including monotonically decreasing RMSE and sustained viewing angles.}
\label{fig:timeseries-20s}
\end{figure*}

\section{Approach}
\label{sec:approach}

This section presents the three components of our framework: dual-path delay architecture (Section~\ref{sec:delay-system}), analytical triangulation covariance (Section~\ref{sec:analytical-cov}), and curriculum learning (Section~\ref{sec:curriculum}). Fig.~\ref{fig:system-architecture} provides a system-level overview.

\subsection{Dual-Path Delay Architecture}
\label{sec:delay-system}

Delays create a fundamental misalignment between the true state and the policy's information~\cite{wang2024addressing}. For covariance-based rewards, this raises a design question: should rewards use privileged clean states~\cite{pmlr-v180-baisero22a} or the same noisy delayed states the policy observes?

As briefly introduced in Section~\ref{sec:dec-pomdp}, our three-stage delay pipeline models distinct physical processes: (1)~staleness from discrete sampling intervals, (2)~latency from computation time or network propagation, and (3)~dropout with probability $p_\text{drop}$. These compose to produce cumulative delays of 50--1000\,ms matching real C-UAS system latencies. Fig.~\ref{fig:delay-pipeline} illustrates the pipeline.

Both paths apply the same delay structure but differ in noise injection. The \emph{noisy path} adds per-field Gaussian noise with standard deviations matching the covariance model in Section~\ref{sec:analytical-cov}, scaled by a curriculum factor $s_\text{noise} \in [0,1]$. The \emph{clean path} delays ground-truth states without noise. Under CTDE, decentralized actors always observe noisy delayed states; the two paths differ only in reward computation. This design isolates whether rewards should optimize the true geometric state or the perception-consistent state available to the policy. Ablation (Section~\ref{sec:experiments}) shows that perception-consistent rewards from the noisy path improve aggregate training-horizon outcomes, while clean-path privileged rewards induce a distinct temporal profile with more stable long-horizon behavior.

For each agent, we maintain ego paths with minimal latency simulating onboard sensors and computations, and other paths with full communication delays, creating the perspective asymmetry central to our formulation.

\subsection{Analytical Triangulation Covariance}
\label{sec:analytical-cov}

We extend Gavin~\etal's angular-noise covariance~\cite{Gavin2024} to derive the closed-form covariance of the triangulated position incorporating multi-source uncertainties: pixel detection noise, pose estimation error, gimbal calibration, and camera intrinsics.

For camera $i$ projecting target $\mathbf{X}$ to pixel $\mathbf{u}_i = \pi(K_i, R_i, \mathbf{t}_i, \mathbf{X})$, linearization yields:
\begin{equation}
\label{eq:linearization}
\delta \mathbf{u}_i \approx J_{\mathbf{X}}^{(i)} \, \delta \mathbf{X} + J_{\boldsymbol{\theta}}^{(i)} \, \delta \boldsymbol{\theta}_i
\end{equation}
where $J_{\mathbf{X}}^{(i)} = \partial \pi_i / \partial \mathbf{X}$ is the Jacobian with respect to target position and $J_{\boldsymbol{\theta}}^{(i)}$ captures sensitivity to nuisance parameters $\boldsymbol{\theta}_i$ (pose, gimbal, intrinsics). Stacking all $C$ cameras gives:
\begin{equation}
\label{eq:stacked-system}
\underbrace{\begin{bmatrix} \delta \mathbf{u}_1 \\ \vdots \\ \delta \mathbf{u}_C \end{bmatrix}}_{\delta \mathbf{u}}
\approx
\underbrace{\begin{bmatrix} J_{\mathbf{X}}^{(1)} \\ \vdots \\ J_{\mathbf{X}}^{(C)} \end{bmatrix}}_{J_{\mathbf{X}}}
\delta \mathbf{X}
+
\underbrace{\begin{bmatrix} J_{\boldsymbol{\theta}}^{(1)} & & \\ & \ddots & \\ & & J_{\boldsymbol{\theta}}^{(C)} \end{bmatrix}}_{J_{\boldsymbol{\theta}}}
\underbrace{\begin{bmatrix} \delta \boldsymbol{\theta}_1 \\ \vdots \\ \delta \boldsymbol{\theta}_C \end{bmatrix}}_{\delta \boldsymbol{\theta}}.
\end{equation}
Accordingly, the effective residual covariance incorporating both pixel noise $\Sigma_\text{pix}$ and stacked nuisance-parameter uncertainty $\Sigma_{\boldsymbol{\theta}}$ is $\Sigma_\text{resid} = \Sigma_\text{pix} + J_{\boldsymbol{\theta}} \Sigma_{\boldsymbol{\theta}} J_{\boldsymbol{\theta}}^\top$. The weighted least-squares information matrix and target position covariance are:
\begin{equation}
\label{eq:info-matrix}
H = J_{\mathbf{X}}^\top \Sigma_\text{resid}^{-1} J_{\mathbf{X}} \in \R^{3 \times 3}
\end{equation}
\begin{equation}
\label{eq:target-cov}
\Sigma_{\mathbf{X}} = H^{-1} = \left( J_{\mathbf{X}}^\top \Sigma_\text{resid}^{-1} J_{\mathbf{X}} \right)^{-1}
\end{equation}

The multi-source $\Sigma_{\boldsymbol{\theta}}^{(i)}$ encodes pixel position uncertainty from object detection, position and orientation error from state estimation, gimbal yaw/pitch calibration error, and focal length/principal point error. This captures why certain configurations yield poor triangulation---near-parallel rays amplify pixel noise along the depth axis, while large baselines mitigate pose uncertainty. The angular-only formulation of~\cite{Gavin2024} captures only one dimension of this error budget.

\subsection{Curriculum Learning}
\label{sec:curriculum}

Direct training with full delays, noise, and moving targets leads to poor convergence because agents must simultaneously learn target visibility, baseline formation, and stale-information handling. We employ a staged curriculum across eight dimensions with linear progress ramps organized in three phases: \emph{Phase~1} (0--60k steps) focuses on controlling the ego motion and the sensor for visual tracking of the stationary target under clean observations; \emph{Phase~2} (20k--100k steps) adds target motion and the triangulation coordination reward; \emph{Phase~3} (80k--200k steps) progressively introduces observation noise, communication delays, detection dropout, and dynamics randomization. This ordering ensures agents learn cooperative geometry under ideal conditions before confronting partial observability. The ablations in Section~\ref{sec:experiments} then isolate the main design choices: recurrent temporal memory with AoI, perception-consistent versus privileged rewards, and multi-source versus angular-only covariance.

\section{Experiments and Results}
\label{sec:experiments}

\subsection{Setup}

All experiments use Isaac Sim 4.5 / Isaac Lab 2.1 with SKRL 1.4.3, running 4096 parallel environments on a single NVIDIA RTX 3090 24GB GPU, AMD Ryzen 9, Ubuntu 20.04, CUDA 12.2, MAPPO under CTDE with parameter sharing and a shared recurrent policy (two-layer MLP encoder mapping 47D observations to 64D, followed by a 64-unit GRU, sequence length 32) to account for delayed observations through hidden-state memory, and the eight-dimensional curriculum for 200k steps ($\approx$819M transitions). We evaluate five conditions at full curriculum difficulty:

\begin{enumerate}
\item \textbf{Perception-consistent reward}: Rewards computed from noisy delayed observations with perception-consistent covariance, AoI, GRU policy, and full delay curriculum.
\item \textbf{Privileged reward}: Identical to the perception-consistent rewards but states are computed from clean delayed states (privileged). Tests Contribution~2.
\item \textbf{No AoI}: Same as perception-consistent reward but AoI removed from observations. Tests Contribution~1.
\item \textbf{Angular-only covariance}: Same as perception-consistent reward but covariance uses only angular noise, matching~\cite{Gavin2024}. Tests Contribution~3.
\item \textbf{MLP (stacked obs)}: Same as perception-consistent reward but GRU replaced by 3-layer MLP (256 units) with frame stacking ($K{=}4$, skip${=}10$, ${\sim}1.2$\,s context). Tests temporal modeling.
\end{enumerate}

Evaluation metrics (Table~\ref{tab:metric-definitions}) are reported with statistics clipped to the 5th--95th percentile over $N > 4{,}000$ episodes per condition.

\subsection{Main Results}

Table~\ref{tab:ablation-results} presents the main results. The noisy-reward (perception-consistent) policy achieves the best localization accuracy ($0.547 \pm 0.217$\,m RMSE) with the lowest collision and track loss rates, while the clean-reward policy attains the highest triangulation validity ($79.1\%$) and visibility ($86.7\%$). Overall, the ablations confirm that each component contributes to at least one key performance metric.
Training dynamics, representative coordination trajectories, and per-timestep metric evolution are shown in Fig.~\ref{fig:training-curves}, Fig.~\ref{fig:trajectories}, and Fig.~\ref{fig:timeseries-20s}, respectively.

\subsection{Analysis}

\textbf{Temporal modeling (C1).} The MLP with stacked observations achieves near-zero triangulation validity ($0.7\%$) despite covering ${\sim}1.2$\,s of temporal context, with the high episode count ($N = 6784$) indicating massive early termination. Under our stochastic delay model, observation timestamps are non-uniformly spaced---the same sequence may contain measurements delayed by 50\,ms and 150\,ms in unpredictable order. Frame stacking provides a fixed temporal window but cannot adapt to this variable-latency structure. The GRU's learned hidden state performs implicit delay compensation, serving a role functionally analogous to Fu~\etal's explicit compensator networks~\cite{fu2025rainbow} but absorbed end-to-end into the policy. Adding AoI to the observation improves RMSE from $0.701$\,m to $0.633$\,m and increases triangulation validity from $68.5\%$ to $79.1\%$ ($+10.6$\,\%p), confirming that explicit staleness metadata provides information that recurrence alone cannot fully recover---AoI offloads temporal bookkeeping, freeing the GRU's limited hidden-state capacity for coordination.

\textbf{Perception-consistent rewards (C2).} Contrary to the privileged-information hypothesis~\cite{pmlr-v180-baisero22a}, perception-consistent rewards outperform clean-state rewards in aggregate metrics: $14\%$ lower RMSE, 27\% fewer track losses, and 38\% fewer collisions. Crucially, both policies observe the same noisy delayed states---they are trained through identical noise---but differ in what they are optimized for. Clean rewards evaluate geometry from ground-truth states, steering the policy gradient toward clean-state optima. Because triangulation covariance is a highly nonlinear geometric function, these optima can lie in noise-sensitive regions of the configuration space: a near-perfect ray intersection angle that appears optimal under clean geometry may border a degenerate zone where small observation perturbations cause near-parallel rays. The clean reward never penalizes this proximity because it does not see the noise. Perception-consistent rewards reshape the optimization landscape itself: the same configuration receives different reward values depending on its sensitivity to the noise actually present, so noise-fragile optima are penalized even when clean-state geometry is favorable. This implicitly teaches the policy to maintain robust margins.

Per-timestep analysis (Fig.~\ref{fig:timeseries-20s}) reveals a complementary \emph{stability--robustness tradeoff}: the clean-reward policy converges to a stable geometric equilibrium---monotonically decreasing RMSE and sustained viewing angles---because the less stochastic reward landscape provides a consistent optimization target with a restoring gradient. The noisy-reward policy, receiving stochastic rewards for the same true configuration due to measurement variation, instead learns reactive coordination that continuously readjusts to current observations. This produces noise-robust behavior but lacks a fixed attractor, resulting in the gradual RMSE increase and visibility decay visible in Fig.~\ref{fig:timeseries-20s}. Representative trajectories (Fig.~\ref{fig:trajectories}) visualize this difference. Neither formulation strictly dominates: the choice between accuracy-oriented perception-consistent rewards and stability-oriented privileged rewards is a deployment-dependent design decision. More broadly, this identifies reward--observation alignment as a design axis complementary to state-reconstruction approaches in delayed-observation RL~\cite{wang2024addressing,fu2025rainbow}.

\textbf{Multi-source covariance (C3).} The angular-only policy achieves $1.546$\,m RMSE with $32.2\%$ validity---a $2.8\times$ RMSE degradation and $-45.9$\,\%p validity drop. Track losses increase from 6.1 to 17.3 ($2.8\times$). The angular-only model captures only one dimension of the uncertainty budget; under realistic conditions with noisy poses and imperfect gimbal calibration, unmodeled noise sources dominate. The multi-source covariance captures the full error budget, guiding the policy toward configurations robust across all uncertainty contributions.

\begin{figure}[!t]
\vspace{2mm}
\centering
\includegraphics[width=\columnwidth]{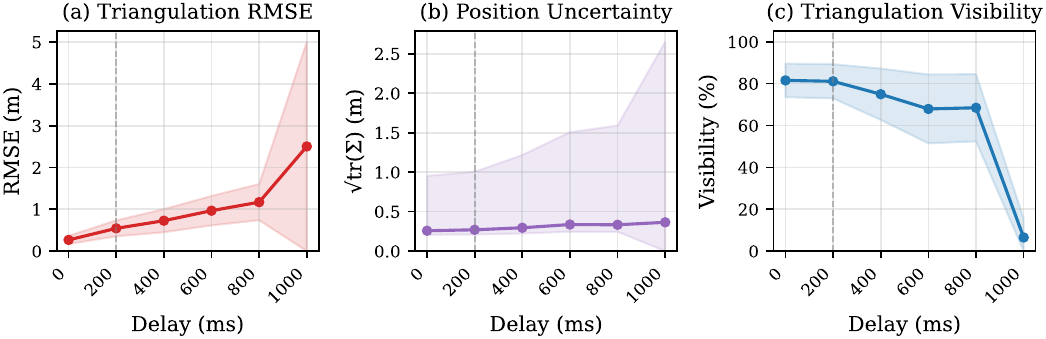}
\caption{Performance envelope across communication delay (0--500\,ms). (a)~RMSE, (b)~position uncertainty $\sqrt{\mathrm{tr}(\Sigma)}$, (c)~visibility. Dashed line indicates training default (100\,ms). The policy degrades gracefully up to ${\sim}200$\,ms before performance drops significantly.}
\label{fig:envelope-delay}

\vspace{2mm}
\includegraphics[width=\columnwidth]{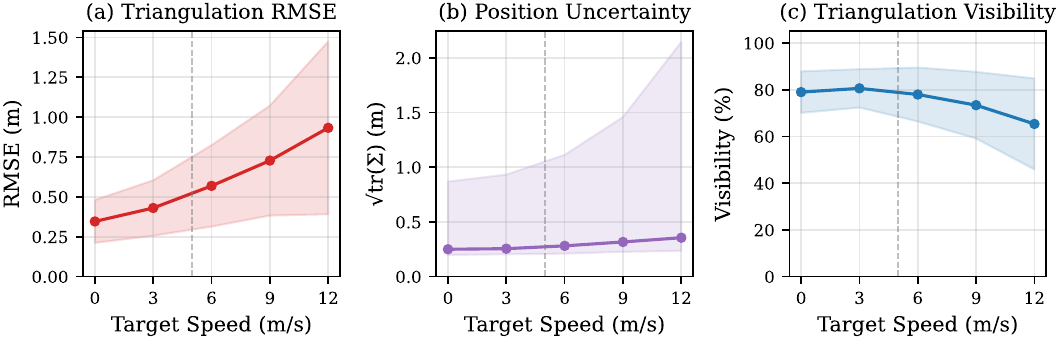}
\caption{Performance envelope across target speed (0--12\,m/s). (a)~RMSE, (b)~position uncertainty $\sqrt{\mathrm{tr}(\Sigma)}$, (c)~visibility. Dashed line indicates training default (5\,m/s). The policy maintains tracking quality across the tested speed range.}
\label{fig:envelope-speed}
\end{figure}

\textbf{Robustness envelopes.} To assess generalization beyond training conditions, we evaluate the perception-consistent policy across communication delays (0--500\,ms, trained at 100\,ms) and target speeds (0--12\,m/s, trained at 5\,m/s). Fig.~\ref{fig:envelope-delay} shows the policy degrades gracefully up to ${\sim}200$\,ms before performance drops significantly. Fig.~\ref{fig:envelope-speed} shows sustained tracking quality across the tested speed range, indicating the learned coordination generalizes beyond the training distribution.

\textbf{Non-stationary delays.} The present delay curriculum samples stationary latency and dropout distributions, but AoI is directly measured from message timestamps and therefore remains available under bursty interference or correlated packet loss. Under such non-stationary channels, AoI would still indicate stale teammate information, while the policy may require additional delay-distribution randomization or online adaptation to match the burst statistics of the deployment environment.

\section{Conclusion}
\label{sec:conclusion}

We presented a delay-aware multi-agent reinforcement learning framework for active visual triangulation to localize an aerial target in Counter-UAS scenarios, unifying AoI-augmented Dec-POMDP formulation, dual-path reward architecture, and multi-source analytical covariance propagation within MAPPO. Systematic ablations validate each contribution: AoI-augmented observations improve triangulation validity by 10.6\,\%p over the ablated baseline; perception-consistent rewards yield $0.547$\,m RMSE with 27\% fewer track losses than privileged rewards, though per-timestep analysis reveals complementary temporal profiles between the two formulations; and multi-source covariance reduces RMSE 2.8-fold compared to angular-only uncertainty modeling. These findings demonstrate that bridging networked control, multi-view geometry, and multi-agent RL enables robust cooperative aerial tracking under realistic communication constraints.

The current study is simulation-based and focuses on the minimal two-agent triangulation case. The delay and noise ranges were chosen to reflect realistic C-UAS perception and communication latencies, but real-flight validation with onboard sensing and wireless networking remains necessary. Scaling beyond two agents is not a fundamental limitation of the estimation objective, the reward formulation, CTDE, or the centralized critic; rather, preliminary three-agent experiments show collision rates rising $9.5\times$, indicating that the present observation and reward do not signal multi-agent collision risk sufficiently. Future work will explore effective pairwise collision encodings for larger teams, hybrid reward strategies that blend privileged and perception-consistent signals, online estimation of deployment noise and delay conditions---which existing delay-aware frameworks~\cite{wang2024addressing,fu2025rainbow} often assume known---uncertainty models that account for measurement staleness, and sim-to-real transfer with real-flight validation.

\section*{Acknowledgment}
This work was supported by the National Research Foundation of Korea (NRF) grant funded by the Korea government (MSIT) (No. NR046525, 2710101596).

\bibliographystyle{IEEEtran}
\bibliography{bib/export}

\end{document}